\documentclass{article}

\usepackage{PRIMEarxiv}

\usepackage{amssymb}
\usepackage{amsmath}
\usepackage{float}
\usepackage{xcolor}
\usepackage{xspace}
\usepackage{booktabs}
\usepackage{makecell}
\usepackage{multirow}
\usepackage{verbatim}
\usepackage{subcaption}
\usepackage[colorlinks=true,linkcolor=blue]{hyperref}
\usepackage[linesnumbered,ruled,vlined]{algorithm2e}
\usepackage{enumerate}
\usepackage[shortlabels]{enumitem}

\usepackage{xurl}
\usepackage{helvet}
\usepackage{etoolbox}
\usepackage{graphicx}
\usepackage{titlesec}
\usepackage{caption}
\usepackage{scrextend}
\usepackage{amsfonts}
\usepackage{bm}
\usepackage{amsmath}
\usepackage{soul}

\usepackage{tikz}
\relax
\usepackage[edges]{forest}
\usetikzlibrary{shadows.blur}

\usepackage{csquotes}
\usepackage{tabularx}
\usepackage{url}
\usepackage{authblk}

\title{The Energy Prediction Smart-Meter Dataset:
Analysis of Previous Competitions and Beyond
}

\author[1,*]{Direnc Pekaslan}
\author[2,*]{Jose Maria Alonso-Moral}
\author[3]{Kasun Bandara}
\author[4,14]{Christoph Bergmeir}
\author[5]{Juan Bernabé-Moreno}
\author[6]{Robert Eigenmann}
\author[7]{Nils Einecke}
\author[8]{Selvi Ergen}
\author[14]{Rakshitha Godahewa}
\author[9]{Hansika Hewamalage}
\author[10]{Jesus Lago}
\author[7]{Steffen Limmer}
\author[7]{Sven Rebhan}
\author[11]{Boris Rabinovich}
\author[9]{Dilini Rajapasksha}
\author[1]{Heda Song}
\author[1]{Christian Wagner}
\author[12]{Wenlong Wu}
\author[13,*]{Luis Magdalena}
\author[4,1,*]{Isaac Triguero}

\affil[1]{School of Computer Science, University of Nottingham, United Kingdom}
\affil[2]{Centro Singular de Investigaci\'{o}n en Tecnolox\'{i}as Intelixentes, Universidade de Santiago de Compostela, Spain}
\affil[3]{School of Computing and Information Systems, University of Melbourne, Australia}
\affil[4]{Department of Computer Science and Artificial Intelligence, University of Granada, Spain}
\affil[5]{IBM Research Europe, Dublin, Ireland}
\affil[6]{E.ON Digital Technology GmbH, Munich, Germany}
\affil[7]{Honda Research Institute Europe GmbH, Offenbach, Germany}
\affil[8]{National Grid ESO, United Kingdom}
\affil[9]{Coles Group, Hawthorn East, VIC, Australia}
\affil[10]{Amazon, Prime \& Marketing Science}
\affil[11]{Amdocs, Israel}
\affil[12]{University of Missouri, Columbia}
\affil[13]{CIS VP TA and Universidad Politécnica de Madrid}
\affil[14]{Department of Data Science and Artificial Intelligence, Monash University, Australia}
\affil[*]{Authors who led the organisation of the competitions and writing of this work. The rest of the authors are ordered alphabetically. Corresponding author: triguero@decsai.ugr.es. }

\begin{document}
\maketitle

\begin{abstract}

This paper presents the real-world smart-meter dataset and offers an analysis of solutions derived from the Energy Prediction Technical Challenges, focusing primarily on two key competitions: the IEEE Computational Intelligence Society (IEEE-CIS) Technical Challenge on Energy Prediction from Smart Meter data in 2020 (named \textit{EP}) and its follow-up challenge at the IEEE International Conference on Fuzzy Systems (FUZZ-IEEE) in 2021 (named as \textit{XEP}). 
These competitions focus on accurate energy consumption forecasting and the importance of interpretability in understanding the underlying factors. 
The challenge aims to predict monthly and yearly estimated consumption for households, addressing the accurate billing problem with limited historical smart meter data. 
The dataset comprises 3,248 smart meters, with varying data availability ranging from a minimum of one month to a year. 
This paper delves into the challenges, solutions and analysing issues related to the provided real-world smart meter data, developing accurate predictions at the household level, and introducing evaluation criteria for assessing interpretability.
Additionally, this paper discusses aspects beyond the competitions:  opportunities for energy disaggregation and pattern detection applications at the household level, significance of communicating energy-driven factors for optimised billing, and emphasising the importance of responsible AI and data privacy considerations. 
These aspects provide insights into the broader implications and potential advancements in energy consumption prediction. 
Overall, these competitions provide a dataset for residential energy research and serve as a catalyst for exploring accurate forecasting, enhancing interpretability, and driving progress towards the discussion of various aspects such as energy disaggregation, demand response programs or behavioural interventions.

\end{abstract}



%

\section{Introduction}
\label{Introduction}
%
%
%
%

\looseness=-1
The energy industry faces a significant obstacle in accurately forecasting energy consumption, as consumption patterns continuously change in given locations \cite{zhao2012review,amasyali2018review,wang2018review}. To address this challenge, predictive modelling techniques and data analytics are commonly employed to forecast future energy usage \cite{mohsenian2010optimal}. While forecasting energy consumption is crucial for the industry as a whole, it becomes even more vital when considering the significant contribution of residential households. Residential energy consumption constitutes a substantial portion of the overall energy demand, making it essential to accurately predict and manage residential energy usage. By leveraging forecasting methods, the industry can not only automate adjustments in energy production to align with demand but also empower residential households to make informed decisions about their energy usage. This leads to improved energy efficiency, cost savings for consumers, and a more sustainable energy ecosystem. Consequently, the ability to forecast energy consumption, particularly in residential households, plays a crucial role in optimising energy utilisation and indirectly contributes to long-term sustainability \cite{limmer2021coordination,jain2014forecasting,stadie2021v2b}.

Smart meters have gained significant popularity and accessibility among households worldwide \cite{wang2018review}. These devices have an essential function of monitoring energy consumption at individual household levels which pave the way from traditional approaches on aggregated consumption data for areas or cities \cite{benzi}. Energy suppliers can leverage this widespread smart meter adoption by offering innovative solutions (e.g., energy forecasting at household levels) that emphasise the benefits of smart meters, to meet customer demands and ensure efficient energy utilisation at the household levels \cite{bayram2017survey}. 

While forecasting techniques utilising smart meter data offer promising potential, the effectiveness of these predictions also relies on the interpretability of the models used \cite{che2012adaptive,Rojat2021} and it is essential for energy companies and customers to understand the underlying factors influencing energy consumption. 
There has been a notable surge in research focusing on interpretability in time series forecasting, highlighting the significance of providing clear explanations for predictions to enable informed decision-making \cite{pinheiro2023short}.


Accordingly, energy consumption prediction competitions have gained significant popularity, attracting researchers, data scientists, and industry professionals. These competitions focus on developing and evaluating the accuracy of energy consumption forecasting models, addressing challenges such as short-term or long-term predictions, energy load forecasting, and renewable energy forecasting \cite{hong2014global, hong2016probabilistic, bergmeir2022comparison, taieb2014gradient}. They not only stimulate innovative solutions but also serve as benchmarks for advancing the state-of-the-art in energy forecasting. A key advantage of these competitions is their utilisation of real-world data, which is typically scarce for research purposes. These real-world data enables participants to develop and test their models in a practical context, enhancing the applicability and reliability of the solutions. Overall, energy consumption prediction competitions play a crucial role in driving advancements in energy forecasting through collaboration, innovation, and the use of real-world data sources.

While most competitions primarily focus on aggregated energy consumption, it is equally important to develop accurate forecasting models at the household level to facilitate tasks such as accurate billing \cite{zhang2018forecasting}. By providing reliable predictions, households can make informed decisions regarding their energy usage, leading to lower energy bills and reduced greenhouse gas emissions. Additionally, accurate household-level energy consumption prediction can inform energy providers and policymakers about expected energy demand, enabling optimised energy distribution strategies. Therefore, accurate household-level energy consumption forecasting plays a pivotal role in achieving significant societal and environmental benefits, particularly in the realm of accurate billing and efficient energy management \cite{Chen2023}.

Researchers participating in energy prediction competitions aim to enhance energy prediction using smart meter data while improving the customer experience. In 2020, the IEEE Computational Intelligence Society (IEEE-CIS) Technical Challenge (named \textit{EP}) on Energy Prediction from Smart Meter data was organised in collaboration with a leading international energy provider, E.ON SE, with a focus on improving energy prediction based on real-world smart meter data~\cite{CIS}. This competition aimed to accurately predict monthly and yearly electricity consumption, addressing numerous challenges inherent to energy forecasting. These challenges encompassed varying data availability, ranging from a minimum of one month to a maximum of one year, diverse seasonality patterns, data gaps due to missing values, and a broad spectrum of household types, among others. These real-world challenges reflect the complexities faced in practice, particularly when dealing with new clients with limited historical data. The solutions developed in these competitions have wider implications for accurate forecasting, benefiting both customers and energy providers.

Considering interpretability is a useful aspect in the realm of machine learning, as it serves to establish trust, enforces accountability, facilitates debugging, ensures regulatory compliance, and offers valuable insights for enhancing the accuracy and effectiveness of the models. 
Therefore, in 2021, at the IEEE International Conference on Fuzzy Systems 
(FUZZ-IEEE), a follow-up competition (named \textit{XEP}) was organised. 
It aimed 
to provide interpretability as well as improve energy prediction accuracy regarding the same smart meter data as in the previous competition. 
While evaluation of interpretability is essential in the context of Trustworthy and Explainable Artificial Intelligence (XAI)
\cite{arrieta2020explainable,Vilone2021,Rudin2022},
it is a challenging task due to the absence of a consensus on how to define and measure interpretability, the subjective nature of human judgement, the complexity of modern models, and the resource-intensive nature of evaluation. Thus, the submitted solutions to the competitions were evaluated by a committee on both accuracy and interpretability. The committee consisted of experts from both academia and industry who assessed the models based on several factors such as model performance, transparency, and ease of understanding. The competition provided a valuable opportunity for researchers and practitioners to showcase their innovative solutions for improving energy prediction and interpretation \cite{FUZZIEEE}.

This paper presents the solutions of two previous competitions, namely EP and XEP, which focused on addressing the challenges related to household-level energy consumption prediction using real-world smart meter data. The paper highlights the following key contributions:

\begin{itemize}
    \item The presentation and discussion of provided solutions: Various solutions were proposed and discussed by participating teams to address the complexities of the smart meter data and to accurately predict household-level energy consumption. These solutions were based on a range of computational intelligence techniques, such as machine learning, deep learning, and statistical models.

    \item The identification and presentation of challenges related to real-world smart meter data: The competition participants identified several challenges associated with real-world smart meter data, such as data sparsity, missing values, and data quality issues. These challenges were thoroughly discussed and analysed to provide insights into the limitations and potential biases in the data.

    \item The establishment of a benchmark problem for researchers: The competitions served as a platform for establishing a benchmark problem that researchers can use to evaluate the performance of their models. This benchmark problem can facilitate the development of more accurate and reliable models for energy consumption prediction.

    \item The development of household-level predictions: The competitions focused on the prediction of energy consumption at the household level, which is essential for optimising energy consumption and reducing energy waste. The proposed models were evaluated based on their ability to accurately predict household-level energy consumption.

    \item The introduction of evaluation criteria for prediction interpretability: The interpretability of machine learning models is crucial for building trust and accountability. Therefore, the competitions introduced evaluation criteria for the interpretability of predictions, which can help researchers to develop more transparent and explainable models.

    \item Pave the way for future research: The competitions open up avenues for future research, allowing the exploration of machine learning techniques to improve the accuracy and interpretability of energy consumption prediction models on real-world data. Furthermore, the competitions extend beyond the initial challenges, facilitating progress in energy disaggregation, demand response programs, and behaviour interventions, all while prioritising responsible AI and data privacy.

\end{itemize}



These competitions provide a valuable platform for 
researchers to tackle real-world challenges and create meaningful impact by developing models that prioritise both accuracy and interpretability. In addition, as real-world data often comes with imperfections and inherent uncertainty, these competitions provide an opportunity for the computational intelligence community to address these challenges and contribute to the understanding of interpretability and uncertainty in the field. By working on these challenging problems with practical applications, these competitions serve as an avenue for researchers to advance the field. The ultimate goal is to develop strategies that enhance the reliability and affordability of energy consumption. Moreover, it is anticipated that the approaches and methods presented in the technical challenge will have broader applicability across various fields facing similar problems of optimal decision-making in the presence of uncertainty.

The paper follows a structured format. In Section II, an overview of the competitions is presented. Sections III, IV, and V delve into the proposed solutions, encompassing pre-processing, prediction, and interpretability approaches, respectively. Each section is accompanied by detailed discussions. In Section VI, the paper highlights the aspects beyond the competitions, emphasising their potential usefulness in energy disaggregation and demand response programs. Section VII provides concludes with a summary and outlines future research directions.


\begin{figure}[t]
    \centering
    \includegraphics[width=0.95\columnwidth]{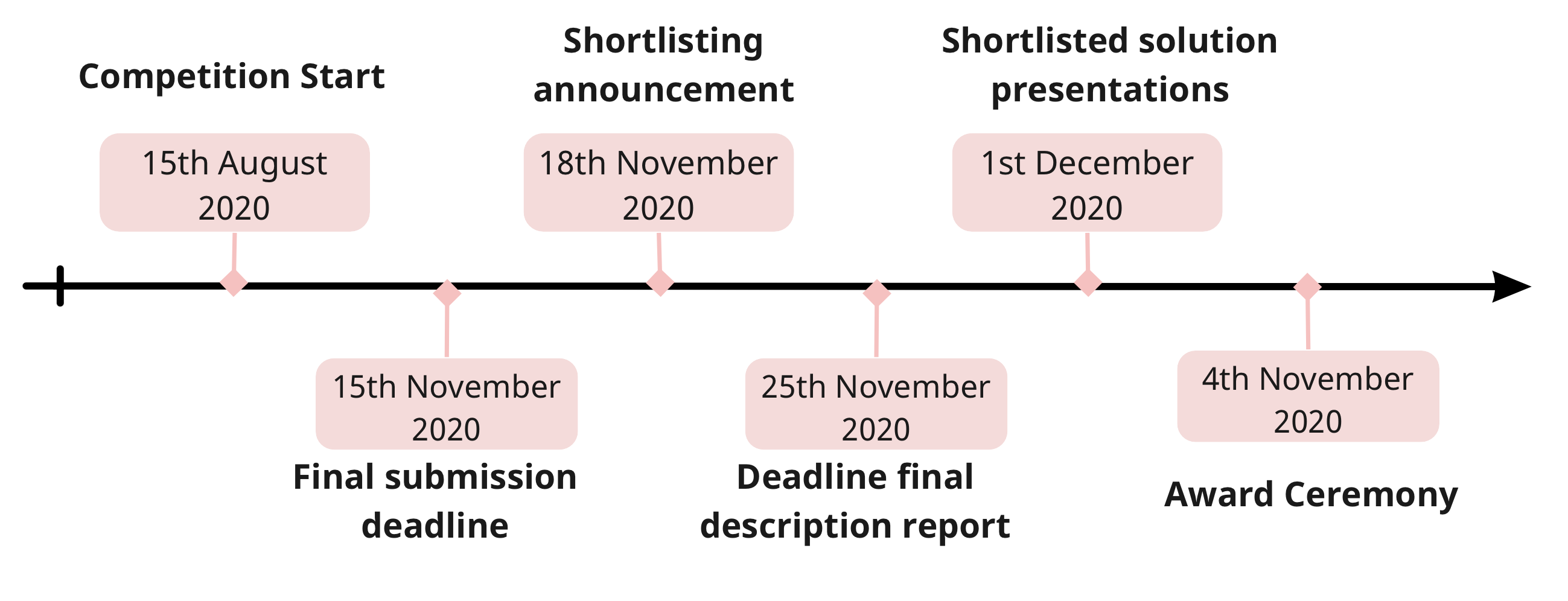}
    \caption{Timeline of the EP Technical Competition in 2020.}
    \label{fig:ieeecisTL}
\end{figure}

\section{Competitions}


As highlighted in the introduction, researchers associated have set out to enhance energy prediction using smart meter data, while also improving the customer experience. To achieve this, a partnership was established with E.ON SE, a leading international energy provider, resulting in the organisation of two competitions, referred to as EP and XEP based on their aspects. 

The EP competition, which took place in 2020 (see the related timeline in Fig.~\ref{fig:ieeecisTL}), was focused on developing accurate predictions of monthly and yearly electricity consumption on a real-world dataset containing several challenges \cite{CIS}. The primary objective was to identify the optimal solutions for energy prediction using smart meters data leading to accurate billing while accounting at household levels. 

\begin{figure}[b]
    \centering
    \includegraphics[width=0.65\columnwidth]{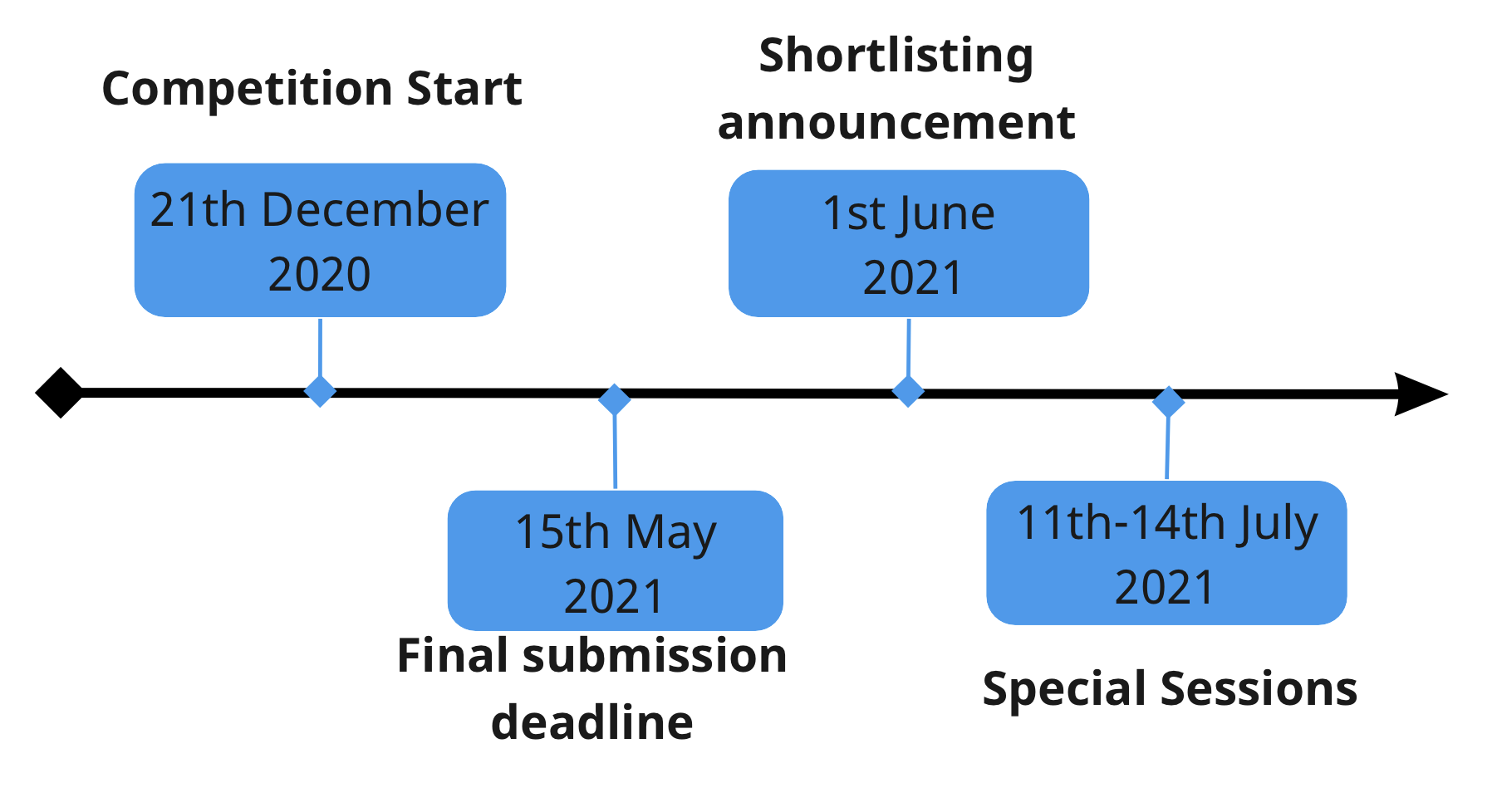}
    \caption{Timeline of the XEP Technical Competition in 2021.}
    \label{fig:fuzzieeeTL}
\end{figure}

Building upon the success of the EP competition, the XEP competition \cite{FUZZIEEE} was held in 2021 (see the related timeline in Fig.~\ref{fig:fuzzieeeTL}). The primary goal was to develop explainable models, accompanied by a narrative explanation in natural language that is easy for customers to understand while maintaining the accuracy of energy prediction. The ability to provide clear explanations of the prediction process is essential to enhance the customer experience and foster greater interpretability with energy usage that can help to build trust, ensure accountability, aid in debugging, ensure regulatory compliance, and provide insights for improving the accuracy and effectiveness of the models. These competitions have contributed to the development of innovative solutions that have the potential to transform the energy industry and improve the sustainability of energy consumption. 









\subsection{Data Description}

The dataset provided comprises half-hourly electricity consumption data obtained from 3,248 household smart meters \cite{CIS,FUZZIEEE}. This dataset covers the period from January 2017 to December 2018. To create a realistic scenario, the prediction task centres on forecasting electricity consumption commencing on January 1, 2018. Notably, each smart meter's data availability varies, ranging from as little as the preceding month (December) to the entire previous year (January to December). This variability accounts for the possibility of customers joining the service at various points throughout the preceding year. Consequently, some customers possess data solely for December 2017, whereas others provide a complete time series spanning from January to December 2017. The main objective is to predict the monthly electricity consumption for the entire year of 2018.

The dataset includes fully anonymous meter IDs and half-hourly consumption data for each smart meter. For the timestamps where data is not available, a nil value is recorded. In addition to the consumption data, the additional information associated with each meter ID is provided as follows:

\begin{itemize}
\item \textbf{Weather data:} One type of additional information is the weather data, which includes daily temperature measurements for the entire year. These data consist of average temperature, minimum temperature, and maximum temperature for each day.

\item \textbf{Additional attributes:} Furthermore, there are additional attributes collected through voluntary surveys. These attributes include information such as dwelling type for 1,702 meters, the number of occupants for 74 meters, and the number of bedrooms for 1,859 meters. It is important to note that not all participants provided complete answers to all survey questions due to the voluntary nature of the survey.
\end{itemize}

\begin{figure}[]
    \centering
    \includegraphics[width=0.7\columnwidth]{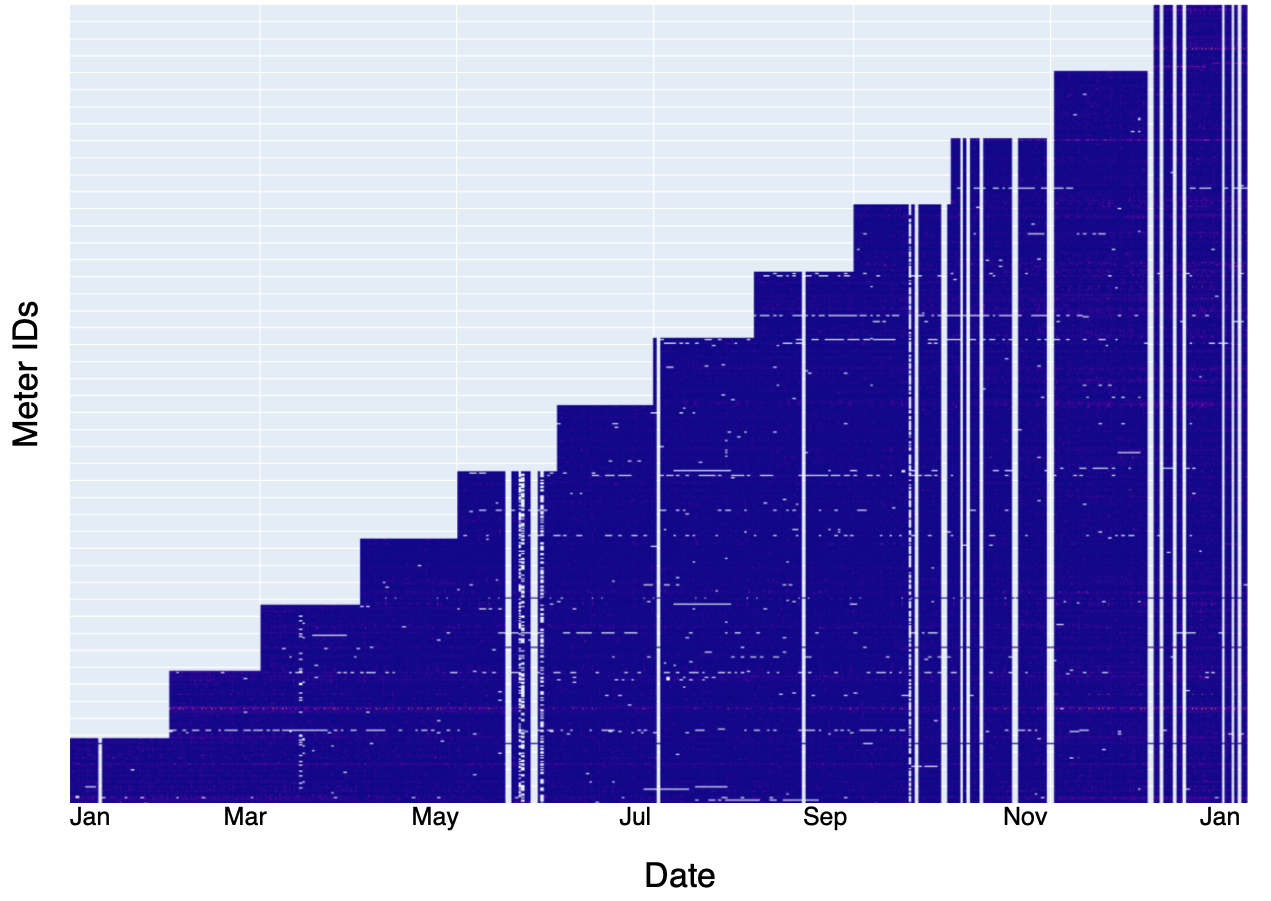}
    \caption{Availability of given data over 2017.}
    \label{fig:avaliablity}
\end{figure}

\subsection{Challenge of the Competitions}

When dealing with real-world data, including smart meter data, it is crucial to acknowledge the range of challenges inherent to such datasets. These challenges should be carefully considered and addressed when developing machine learning systems for prediction purposes.

\begin{itemize}

    \item \textbf{Data quality and completeness}: The presence of inconsistencies, such as missing values or variations in data quality, across different smart meters can pose challenges for accurate energy consumption forecasting at the household level. Addressing data inconsistencies and implementing appropriate data imputation techniques are essential to ensure reliable predictions.

    \item \textbf{Seasonality and temporal patterns}: Energy consumption patterns exhibit seasonality and temporal dependencies. However, the presence of missing or inconsistent seasonal data can make it difficult to identify and predict these patterns accurately. Incorporating techniques that account for seasonality and capture temporal dynamics can improve the forecasting accuracy.

    \item \textbf{Variability in data availability}: The availability of smart meter data can vary among households, with some households registering later than others. This variability in data availability poses a challenge for prediction models, as the absence of historical data for certain households can impact the accuracy of energy consumption forecasts.

    \item \textbf{Limited availability of external data sources}: External data sources, such as weather and economic data, can provide valuable context for understanding energy consumption patterns. However, the lack of such contextual data or inconsistencies in their availability can limit the accuracy of forecasting models. Exploring alternative data sources or leveraging techniques to handle missing external data can help overcome this challenge.
    
\end{itemize}

\begin{figure}[t]
    \centering
    \includegraphics[width=0.65\columnwidth]{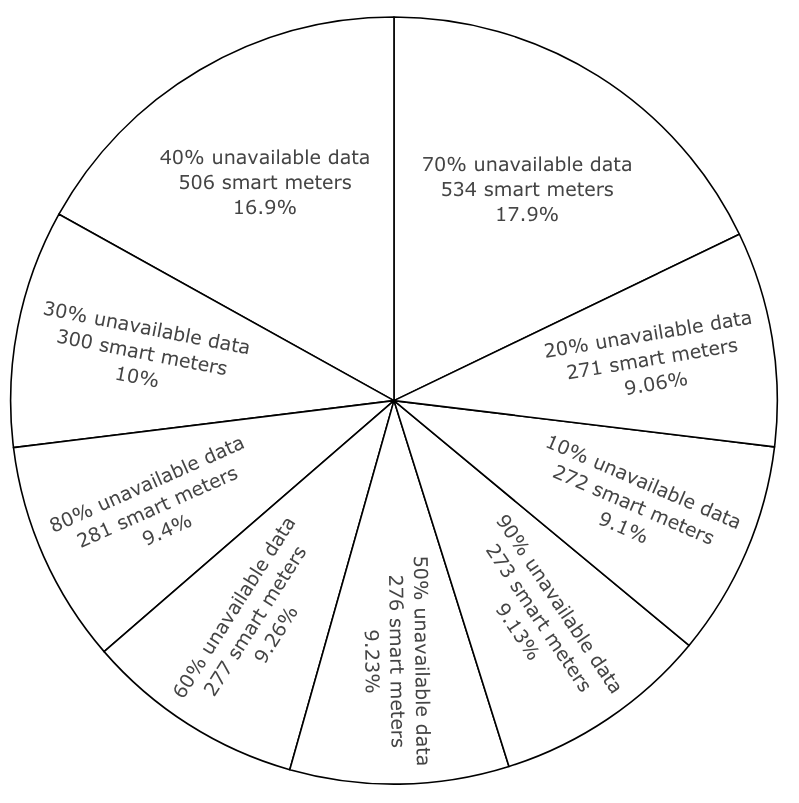}
    \caption{Number of smart meters and corresponding missing data percentage.}
    \label{fig:prct}
\end{figure}

\subsubsection{Data Quality and Availability} The availability and completeness of data play a crucial role in the success of any data analysis or modelling project. In the context of energy forecasting, having access to complete and accurate data is particularly important due to the complex nature of energy consumption patterns. As shown in Fig. \ref{fig:avaliablity}, data availability can vary significantly across different time periods and meter IDs. This variability can present a significant challenge when developing predictive models, as missing or incomplete data can lead to inaccurate or biased predictions.

The pie chart in Fig. \ref{fig:prct} further highlights the severity of the availability issue, indicating that a significant proportion of the dataset has a high level of unavailable values. For instance, 
the data from 534 smart meters exhibits 70\% unavailability, meaning that a significant portion of the data for these meters is missing or unavailable.  Such levels of unavailable data could lead to inaccurate and unreliable predictions, making it difficult for energy companies to make informed decisions and manage energy consumption effectively. 

\subsubsection{Additional Data} In order to obtain a more comprehensive dataset, a survey was conducted to gather additional information about households such as the number of occupants, number of bedrooms, and dwelling types. However, the results of the survey were limited as only 2144 responses were received out of 3248 households. Furthermore, not all respondents answered each question, resulting in missing data for some variables. For example, the question regarding the dwelling type was only answered by 1702 respondents, while 442 respondents did not provide an answer. Similarly, the question about the number of occupants was not answered by the majority of respondents, with only 74 answers received. The details of the survey questions and the number of given answers can be seen in Table \ref{externalDataSurvey}. This incomplete dataset highlights the challenges of obtaining accurate and complete information about households, which is helpful for accurately predicting household-level energy consumption.

\begin{table}[ht]
\caption{Details about the conducted survey for external data collection and the number of responses.}
\label{externalDataSurvey}
\begin{tabularx}{\columnwidth}{|X|X|c|}
\hline
\textbf{Questions} & \textbf{Meaning} & \textbf{\# Responses} \\ \hline
meter\_id            & Anonymised smart meter ID & 1859 \\ \hline
Dwelling type        & Type of building  & 1702 \\ \hline
Number of occupants  & Number of people living in the dwelling  & 74 \\ \hline
Number of bedrooms   & Number of bedrooms in the property  & 1859 \\ \hline
Heating fuel         & Type of fuel used for space heating & 78 \\ \hline
Hot water fuel       & Type of fuel used for heating water & 76 \\ \hline
Boiler age           & Age of the boiler (new or old)  & 70 \\ \hline
Loft insulation      & Presence of insulation in the loft  & 70 \\ \hline
Wall insulation      & Presence of insulation in the walls  & 73 \\ \hline
Heating temperature  & Typical indoor temperature set for heating  & 76 \\ \hline
Efficient lighting   & Percentage of lighting that is energy-efficient  & 70 \\ \hline
Dishwasher           & Number of dishwashers in the dwelling  & 70 \\ \hline
Freezer              & Number of freezers  & 73 \\ \hline
Fridge freezer       & Number of combined fridge-freezers  & 76 \\ \hline
Refrigerator         & Number of refrigerators  & 76 \\ \hline
Tumble dryer         & Number of tumble dryers  & 72 \\ \hline
Washing machine      & Number of washing machines  & 70 \\ \hline
Game console         & Number of game consoles  & 70 \\ \hline
Laptop               & Number of laptops  & 69 \\ \hline
PC                   & Number of personal computers  & 70 \\ \hline
Router               & Number of internet routers & 70 \\ \hline
Set top box          & Number of set-top boxes for television  & 75 \\ \hline
Tablet               & Number of tablet computers  & 69 \\ \hline
TV                   & Number of televisions  & 70 \\ \hline
\end{tabularx}
\end{table}

\subsubsection{Lack of Pattern} As exemplified above, challenges exist in real-world datasets. Such as 
challenges includes the presence of various types of seasonality, which can have a significant impact on energy consumption patterns. Additionally, incomplete data due to missing values, as well as the use of external data sources to handle seasonal effects, further complicate the prediction process. Another challenge arises from the diversity of household types, such as differences in energy usage patterns between families and single households or between older and newer homes. Furthermore, there may be variations in the registration time of different households or a lack of sufficient information about specific households, which can further complicate the prediction process. In addition, due to the lack of data, models need to be trained using short periods of data to make long-term predictions or predict end-of-year bills which pose another challenge in predictions. Addressing these challenges requires the implementation of advanced machine learning techniques and innovative data processing strategies to generate accurate predictions. 
The need for interpretability in these models also presents a well-known trade-off between accuracy and interpretability, which requires skilled management to balance these competing demands effectively. Overall, overcoming these challenges is critical to the effective use of machine learning in the energy industry and optimising energy consumption and sustainability.


\subsection{Evaluation}

To provide a starting point for the energy consumption prediction task, the competitions adopted a straightforward and naive approach by calculating the average monthly consumption for each smart meter. The computed average was then utilised as the prediction for all months in the subsequent year. This simplistic method served as a baseline, offering a reference point for comparison with more sophisticated models from participants. 

Furthermore, to complement this baseline approach at the XEP competition, organisers developed an explainable fuzzy system that enabled the extraction of baseline explanations in natural language. 
To do so, the open-source software GUAJE\footnote{\url{https://demos.citius.usc.es/guajeonline/}}~\cite{pancho2013} is used to implement the methodology for building self-explaining fuzzy systems~\cite{alonso2021}.
Namely, a Mamdani Type-1 fuzzy system is considered which is made up of IF-THEN conjunctive rules. More precisely, rules were learned with the well-known Wang and Mendel algorithm~\cite{WM1992}. Since this fuzzy system was taken as baseline for generating explanations in natural language, we did not apply any of the advance tools provided by GUAJE for optimising its interpretability-accuracy trade-off.
As a result, GUAJE produced rules like ``IF A and B THEN C" that were automatically translated into linguistic explanations like ``The estimation is C because A and B".
Following this procedure, the baseline generates
explanations such as: 

\begin{displayquote}
``The estimation of your energy consumption for next year is low because your average monthly consumption this year has been high", \end{displayquote}
or 

\begin{displayquote} ``In January, your energy consumption will be low because your average monthly consumption this year has been low".
\end{displayquote}

By associating these explanations with corresponding numerical values, the naive approach and the fuzzy system serve as essential starting points to participants.

In the following, we first provide the accuracy evaluation measures and second the explainability assessment procedures are detailed. 

\subsubsection{Accuracy Evaluation}


Making predictions for smart meters for which having a little historical consumption data is usually a challenging task. In addition to that, an accurate monthly consumption prediction is also very valuable for Energy Trading teams, who need to be able to predict with some accuracy how much electricity to buy on the energy market. Thus, 
we measured accuracy of submissions as follows:

\textbf{Total Year Consumption:} The predicted total year consumption is computed as the sum of all the predictions for each individual month. Then, for each smart meter, a Relative Absolute Error (rAE) is calculated between the predicted total year consumption values and real values, as follows:

\begin{equation}
    year_{rAE}=\frac{\frac{1}{N}{} \sum_{k}^{N}|y_k-t_k|}{\frac{1}{N}{} \sum_{k}^{N}|t_k-\bar{t}|}
\end{equation}
where $\bar{t}=\frac{1}{N} \sum_{k}^{N}|t_k|$, N is the total number of smart meters, $y_k$ is the predicted total year 
consumption of the k-th meter, $t_k$ is the true total year consumption of the k-th meter.

\textbf{Monthly Consumption:} For each smart meter, the rAE between the predicted monthly consumption values and the real values is calculated.

\begin{equation}
    month_{rAE}=\frac{1}{N}\sum_{k}^{N}\frac{\frac{1}{12}\sum_{i}^{12}|y_k^i-t_k^i| }{\frac{1}{12}\sum_{i}^{12}|t_k^i-\bar{t_k}|}
\end{equation}
where $\bar{t}=\frac{1}{12} \sum_{i}^{12}|t_k^i|$, $y_k$ is the predicted monthly consumption of the k-th meter, $y_k=[y_k^1,y_k^2,...,y_k^{12}]$ includes predicted total month consumption for 12 months, $t_k$ is the true monthly consumption of the k-th meter, $t_k=[t_k^1,t_k^2,...,t_k^{12}]$ includes true total month consumption for 12 months.

Overall, 
both metrics are considered equally important and aggregated as:

\begin{equation}
    total_{rAE}=\frac{1}{2}year_{rAE} + \frac{1}{2}month_{rAE}
\end{equation}

\subsubsection{Explanation Evaluation}






A panel of five experts is in charge of assessing goodness and effectiveness of narrative explanations associated to predictions of both total year and monthly consumption, regarding a selection of 
smart meters from the test set. 
These experts were all members of the IEEE-CIS Task Force on Explainable Fuzzy Systems.

The panel considers the structural and semantic complexity of solutions, regarding accountability, clarity, naturalness, completeness, and length of explanations in an off-line blind evaluation procedure. 
Each criterion in the list below is evaluated with a 5-point Likert scale as in: 1) Strongly Disagree; 2) somewhat disagree; 3) neutral; 4) somewhat agree; 5) Strongly Agree.


\begin{itemize}
    \item C1. ``The explanation of the prediction is formally correct (i.e., readable, fluent and grammatically correct)". 

    \item C2. ``From the explanation, the expert understands how the prediction is automatically made".

    \item C3. ``From the explanation, the expert understands the cause and effect of the prediction".

    \item C4. ``The explanation comprises factual and counterfactual pieces of information".

    \item C5. ``The explanation has sufficient detail".

    \item C6. ``The explanation is complete".

    \item C7. ``The explanation is useful to my goals".

    \item C8. ``The explanation shows how accurate the prediction is". 

    \item C9. ``The explanation allows the expert to judge when should the prediction be trusted or not trusted".
    
\end{itemize}

In addition, a global evaluation (C10. ``The explanation of how the predictor works is satisfying") was required to be answered in a 5-point Likert Scale as: 1) Very Dissatisfied; 2) Dissatisfied; 3) Neither Satisfied Nor Dissatisfied; 4) Satisfied; 5) Very Satisfied.

All submissions were shortlisted with respect to accuracy and the three top-ranked methods were nominated as finalists.
To assess the performance and interpretability of the three finalists, a dedicated website\footnote{\url{https://tec.citius.usc.es/xaicompetition}} was developed,
serving as a platform to present the results and gather expert
evaluations for the 10 criteria previously introduced.
Ten smart meters from the test set were 
chosen for the evaluation, and their yearly and monthly predictions were displayed to the experts, specifically focusing on the months of February, May, and December. To ensure a focused and manageable evaluation process, a deliberate decision was made to limit the number of smart meters to ten and select only three specific months for the monthly prediction explanations. This approach was adopted to avoid overwhelming the experts with an excessive amount of information, allowing for a more in-depth analysis of the selected cases and facilitating a comprehensive assessment of the three methods under evaluation in terms of performance and interpretability.
It is worth noting that the selected test cases cover a variety of challenging scenarios which were of interest for the organisers of the competition, i.e., EP and E.ON SE.

\begin{figure}[htb]
    \centering
    \includegraphics[width=1\columnwidth]{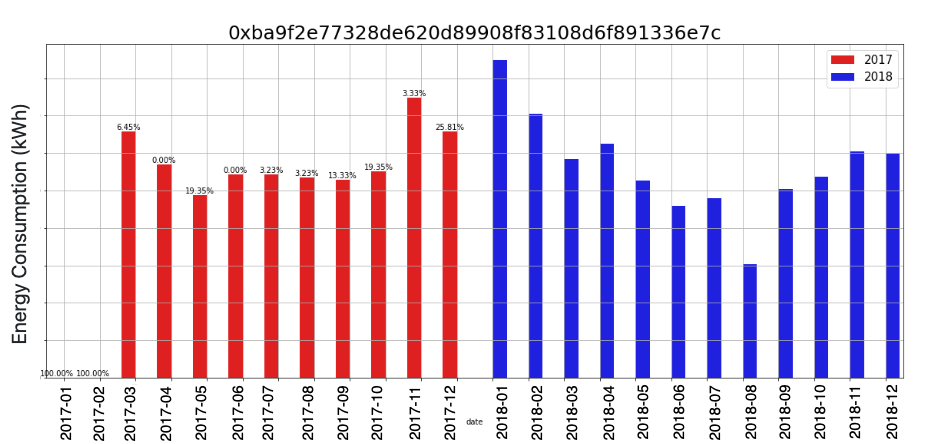}
    \caption{Actual data visualisation for energy consumption patterns of a randomly selected smart meter. Monthly energy consumption data, providing an overview of consumption trends over time, displaying 2017 red and 2018 blue bars.}
    \label{fig:smartmeter}
\end{figure}

For each given prediction, explanations were provided to the panel of experts, along with visualisations of the corresponding actual data (see Fig. \ref{fig:smartmeter}) and any additional information available for the selected smart meters. The evaluation process was conducted anonymously, by filling in the given online form, to ensure impartiality. Based on the criteria summarised above, the experts 
assessed the goodness of explanations associated to the ten smart meters for each of the three finalist methods, considering the prediction performance-explanation and real data. This comprehensive evaluation facilitated an in-depth understanding of the strengths and limitations of the methods, aiding in the selection of the most effective and interpretable approach for energy consumption prediction.

\begin{figure}[]
    \centering
    \includegraphics[width=1\columnwidth]{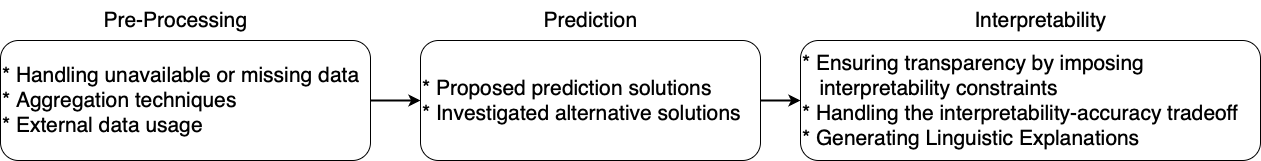}
    \caption{Workflow illustrating the investigation of pre-processing, prediction, and interpretability techniques, highlighting the approach adopted for each stage of the analysis.}
    \label{fig:workflow}
\end{figure}

At the end, finalists are ranked based on an aggregated score of prediction and explanation performance. The panel considered both error rates and explanations as a whole. The total $rAE$ weights both indicators equally. Scores of C1 – C10 are averaged for all given explanations. 
In addition, each finalist submitted a report describing their method. Accordingly, the panel also has to carry out a qualitative review of the description of the methodology applied by each finalist in terms of: 
\begin{enumerate}
    \item Clarity and justification of the proposed methodology: ``Does it appropriately use Computational Intelligence techniques?"

    \item Robustness and potential generalisation of the proposed approach, regarding both prediction and automatic narrative explanation generation: ``Is this a heavily ad-hoc solution?"

    \item Experimental methodology and parameter tuning: ``Have the participants tuned their parameters against the test dataset?"
    
\end{enumerate}

The overall assessment 
is considered together with the aggregated scores and qualitative reviewer feedback to determine the winners.

\subsection{Technical and Scientific Committee }

In order to ensure the success of the EP and XEP competitions, a team of dedicated organisers and scientific committee members were assembled to oversee the technical aspects and evaluate the results. The organisers for both competitions included leading experts who were responsible for establishing the competition rules, setting up the infrastructure for data management, and ensuring each competition was executed smoothly.

The scientific committee members, on the other hand, were responsible for evaluating the results submitted by the competition participants. They were chosen based on their expertise in machine learning, data analysis, and energy prediction. The committee members played a critical role in assessing the accuracy of the predictions and the quality of the explanations provided in the self-explaining models developed for the XEP competition.

The combined efforts of the organisers and scientific committee members were crucial in ensuring the success of both competitions. Through their expertise and commitment to excellence, they helped to foster a collaborative environment that encouraged innovation and creativity. By providing a platform for researchers and practitioners to showcase their skills and solutions, the competitions have contributed to advancing the field of energy prediction and smart meter data analysis. 

\begin{figure*}[t]
    \centering
    \includegraphics[width=1\columnwidth]{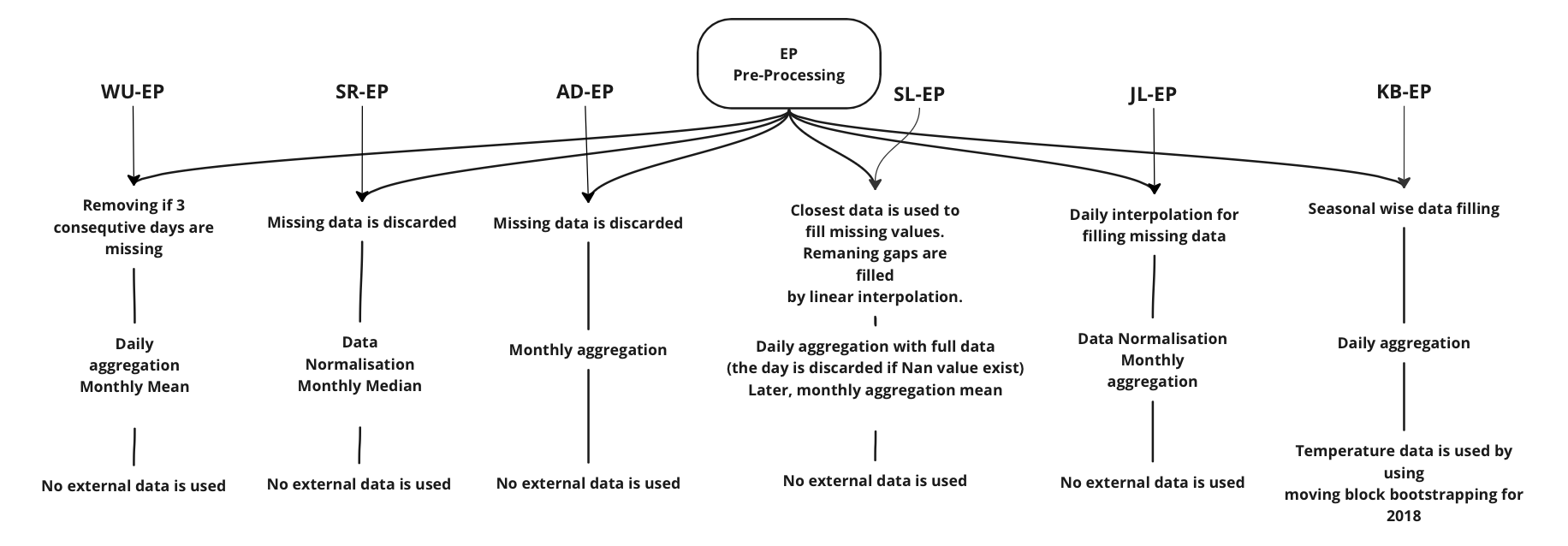}
    \caption{Pre-processing stage of the workflow, showcasing the EP strategies employed to address missing/unavailable data filling, data aggregation, and the incorporation of external data for enhancing the quality and completeness of the dataset.}
    \label{fig:TaxPreProcessingCIS}
\end{figure*}

In the upcoming sections, we will delve into the solutions proposed by the shortlisted participants of the EP and XEP competitions. As illustrated in Fig. \ref{fig:workflow}, to provide a comprehensive overview, we will first examine each shortlisted participant's pre-processing approach, which they employed to clean and prepare the raw smart meter data for analysis. Next, we will delve into their prediction approaches, which were tailored to the specific requirements of each competition. For the EP and XEP competitions, we will examine the prediction models developed by each shortlisted participant and evaluate their performance based on accuracy and efficiency. Additionally, for the XEP competition, we will examine the interpretability of reported solutions which were aimed at providing insight into how the models arrived at their predictions. Through this comprehensive analysis, we shed light on the cutting-edge techniques and innovative approaches being used to tackle the complex challenges of energy prediction and smart meter data analysis.

\section{Pre-processing}

In this subsection, we will take a closer look at the pre-processing methodologies employed by the participants in both the EP and XEP competitions. Specifically, as shown in Fig. \ref{fig:workflow}, we will examine their approaches to data aggregation, filling missing values, and the use of external data sources.

\subsection{EP Pre-Processing Solutions}

Regarding the EP competition, it can be observed that the majority of the finalists, that is 4 out of 6, utilised monthly aggregation, with the remaining contestants adopting daily aggregation as part of the pre-processing procedure. Due to variations in data availability over the course of the year, multiple methods of data imputation were employed; however, the predominant approach involved eliminating any data that was not available. Although certain external data was made available for the prediction task, only one finalist out of the six incorporated external weather data in their analysis. 

\subsubsection{Wenlong Wu's Solution (WU-EP)}

One pre-processing technique involves the removal of any 0-reading or missing data that is continued for a 3-day window. For the remained, the data is aggregated on a daily basis, and a smoothing operation is applied to obtain the monthly mean. To further process the data, cyclical features that encode the sine and cosine values of the month are used to map the data.

\subsubsection{Sven Rebhan's and Nils Eineck's Solution (SR-EP)}
Another method involves discarding any missing data and normalising the meter readings on an individual basis for each month. The normalised meter readings for each month are then subjected to a median operation and provided as input to the model. This approach does not utilise any weather data or additional data pertaining to the meter's environment.

\subsubsection{Alexander Dokumentov's and Fedor Dokumentov's Solution (AD-EP)}

Another approach used by a contestant involves the exclusion of missing values and the calculation of the average daily consumption for each month. The Box-Cox transformation method is applied for normalisation purposes, and the missing values are subsequently imputed at a monthly level using the softImpute package in R, as described in \cite{softImpute, mazumder2010spectral}. The smart meters are then divided into 12 groups based on the amount of missing data, with Group 
$G_0$ having no missing observations and Group 
$G_{11}$ having the first 11 months with missing observations. In this method, Group $G_0$ is utilised in the prediction stage.

\subsubsection{Steffen Limmer's Solution (SL-EP)}

In this particular case, missing values are filled with the first available corresponding value from the previous day, the subsequent day, from two days before and two days after (in this order). Remaining missing values of up to two consecutive time steps are filled via linear interpolation Subsequently, the consumption data measured at half-hourly intervals is re-sampled to the level of daily consumption. During the re-sampling process, any days that have at least one missing half-hourly consumption are treated as missing data. In addition, monthly aggregation is performed with a caveat that if more than 5 days' worth of data is missing, the consumption is deemed unknown. Following this procedure, 270 meters are left with no missing values, and the consumption is calculated and expressed as a fraction or percentage over each month.

\subsubsection{Jesus Lago's Solution (JL-EP)}

The data is subjected to linear interpolation across days, and then aggregated to a monthly level. The monthly consumption for each meter is then normalised by dividing it by the yearly consumption, and the relative consumption fraction is calculated. To overcome the issue of regularisation and find similar sign-up meters, clustering is employed, with the elbow method utilised to determine the optimal cluster number through K-means clustering. Signed-up meters are then grouped together based on the starting months.

\subsubsection{Kasun Bandara's, Hansika Hewamalage's, Rakshitha Godahewa's Solution (KB-EP)}

The half-hourly data is aggregated at the daily level, and any gaps in the data are filled using a seasonal-based approach, which involves calculating the weekly median energy consumption for each household. In addition, the research team employed an alternative technique compared to other competitors by using external data sources. This involved utilising the simulation moving block bootstrapping method to obtain temperature data from 2018.

Having now reviewed the pre-processing methodologies employed in the EP competition (see a visual summary in Fig.\ref{fig:TaxPreProcessingCIS}), we turn our attention to the XEP competition (see Fig.\ref{fig:TaxPreProcessingFUZZ}). In order to prepare the data for analysis, participants in the XEP competition employed a range of approaches.


\subsection{XEP Pre-Processing Solutions}

In the XEP competition, the majority of the finalists, specifically 3 out of 4, employed daily aggregation as part of their data pre-processing strategy. This approach involves the consolidation of half-hourly consumption data into daily values, allowing for a more simplified and manageable dataset. However, one finalist chose to utilise the data in its original half-hourly sampling rate, foregoing the data aggregation step altogether. This alternative approach may be more computationally expensive and require more advanced modelling techniques, but may also provide a more detailed and granular understanding of the consumption patterns.

\subsubsection{Kasun Bandara's and Boris Rabinovich's Solution (KB-XEP)}

The contestants applied a pre-processing methodology similar to the one used in the previous EP competition (see KB-EP in Fig.\ref{fig:TaxPreProcessingCIS}). The data is first aggregated to the daily level, and then missing data are filled through a seasonal-based approach, which consists of utilising the weekly median energy consumption for each household. Additionally, as in the prior competition, temperature data is generated for 2018 through the simulation moving block bootstrapping methodology. This technique is utilised to obtain external data to help inform and refine the energy consumption prediction models.

\subsubsection{Dilini Rajapaksha's and Christoph Bergmeir's Solution (DR-XEP)} 

The half-hourly data is aggregated at a daily level. Apart from this, the simulation approach is employed to depict future weather data. This is accomplished by using the bootstrap method, which is applied to the temperature dataset to produce estimates for the next year's weather conditions. The resulting simulated data is then utilised as an external data source in the model to improve the prediction accuracy.

\subsubsection{Young-ho Cho, Byoungryul Oh,	Juhwan Park	Sooyeon Kim, Duehee Lee's Solution (YC-XEP)} 

To align with the daily frequency of the temperature data, daily aggregations are performed on the power consumption data. The missing temperature data are filled using a method known as latent factor-based collaborative filtering (CF). In addition to temperature data, information about the day of the week is also extracted and integrated to the dataset which will be used in the prediction stage of the study.

\begin{figure}[t]
    \centering
    \includegraphics[width=0.8\columnwidth]{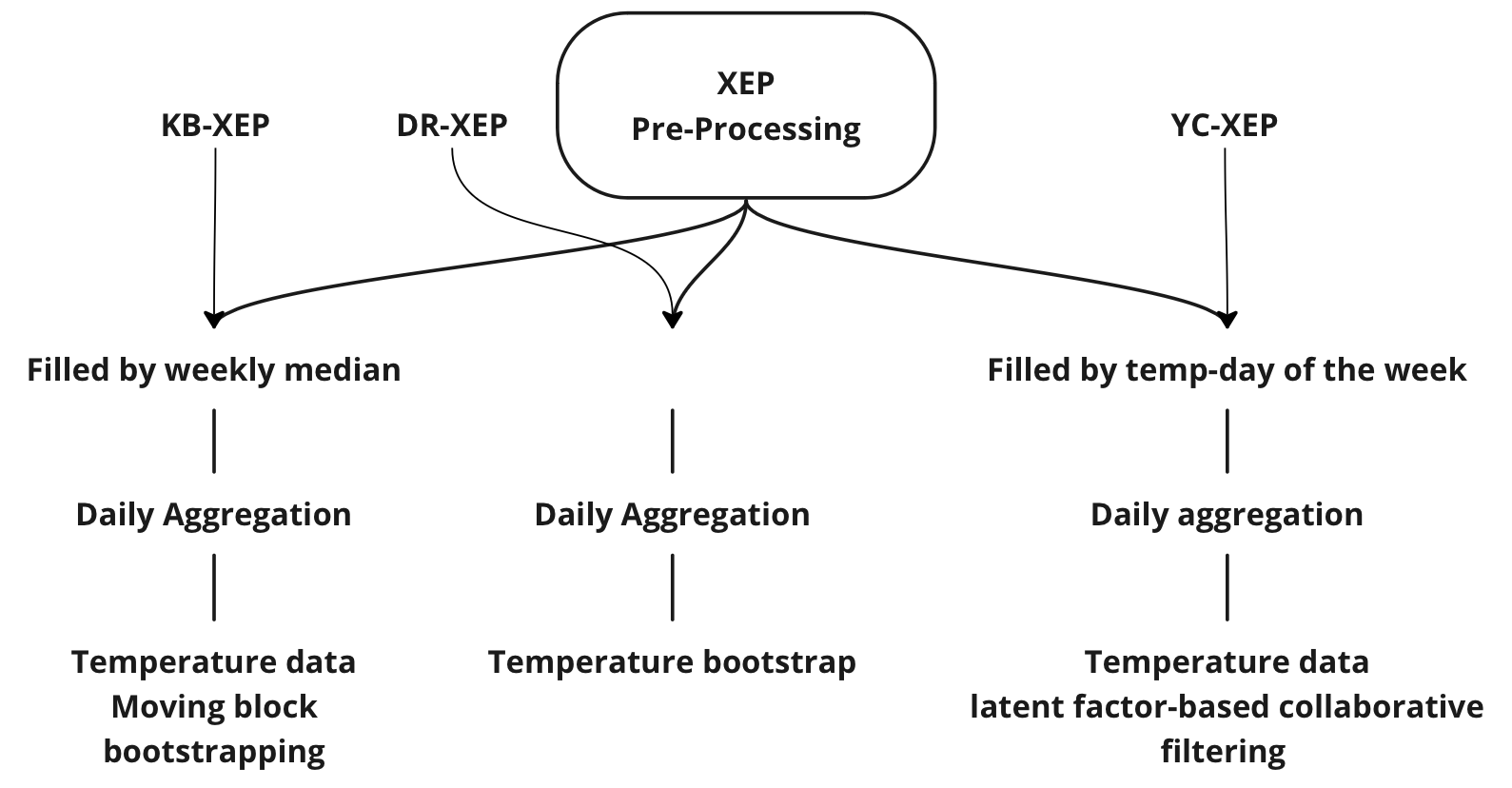}
    \caption{Pre-processing stage of the workflow, showcasing the XEP strategies employed to address missing/unavailable data filling, data aggregation, and the incorporation of external data for enhancing the quality and completeness of the dataset.}
    \label{fig:TaxPreProcessingFUZZ}
\end{figure}



\subsection{Overall Pre-Processing Approaches}

Let us now summarise the main conclusions drawn from pre-processing solutions in both competitions. On the one hand, in the EP competition, six finalists used different techniques for data aggregation and filling missing values. Four of the finalists used monthly aggregation, while the rest used daily aggregation as shown in Fig. \ref{fig:TaxPreProcessingCIS}. Data imputation methods varied due to differences in data availability, but most of the contestants discarded any missing data. Only one finalist used external weather data in 
the analysis. Methods of data filling ranged from discarding to using nearest available values or linear interpolation. In some cases, the consumption data was re-sampled to a daily level or normalised by dividing it by the yearly consumption. Clustering and K-means were used to group signed-up meters. One finalist used an alternative technique involving external data sources, specifically temperature data from 2018, which was obtained using a simulation moving block bootstrapping method.

On the other hand, in the XEP competition, as shown in Fig. \ref{fig:TaxPreProcessingFUZZ}, the majority of finalists used daily aggregation to consolidate half-hourly consumption data into a more manageable dataset. However, one finalist used the data in its original half-hourly sampling rate to provide a more detailed understanding of consumption patterns. In terms of data imputing, some finalists used seasonal-based approaches or latent factor-based collaborative filtering to impute missing data, while one finalist used a gradual extension search method. In terms of external data usage, most finalists used simulated temperature data, while one finalist used daily temperature data extended to 48 observations to ensure consistency with power consumption data.

Finally, regarding 
the available additional dataset, such as the type of dwelling and number of rooms, it was not utilised or was excluded due to data insufficiency or inconsistency, in both competitions. While data aggregation was commonly employed, monthly aggregation was used in the EP competition, 
and daily aggregation was mostly used in the XEP competition. In the EP competition, most participants removed the missing values, whereas in the XEP competition, several methods were used to impute the missing data, and simulated temperature data for the upcoming year was generated.


\section{Prediction}

In this section, we will delve into the prediction solutions proposed by the shortlisted participants in both the EP and XEP competitions. As pointed out in Fig. \ref{fig:workflow}, we will provide a detailed overview of their prediction models where the information is available, highlighting their unique features and strengths. Our aim is to provide readers with an understanding of the diverse approaches adopted by participants in tackling the challenges posed by the real-world datasets provided in both competitions.
Fig. \ref{fig:TaxPredictionCIS} summarises the EP prediction solutions.
Fig. \ref{fig:TaxPredictionFUZZ} summarises the XEP prediction solutions.

\subsection{EP Prediction Solutions}

\subsubsection{WU-EP} 
\begin{figure}[]
    \centering
    \includegraphics[width=0.5\columnwidth]{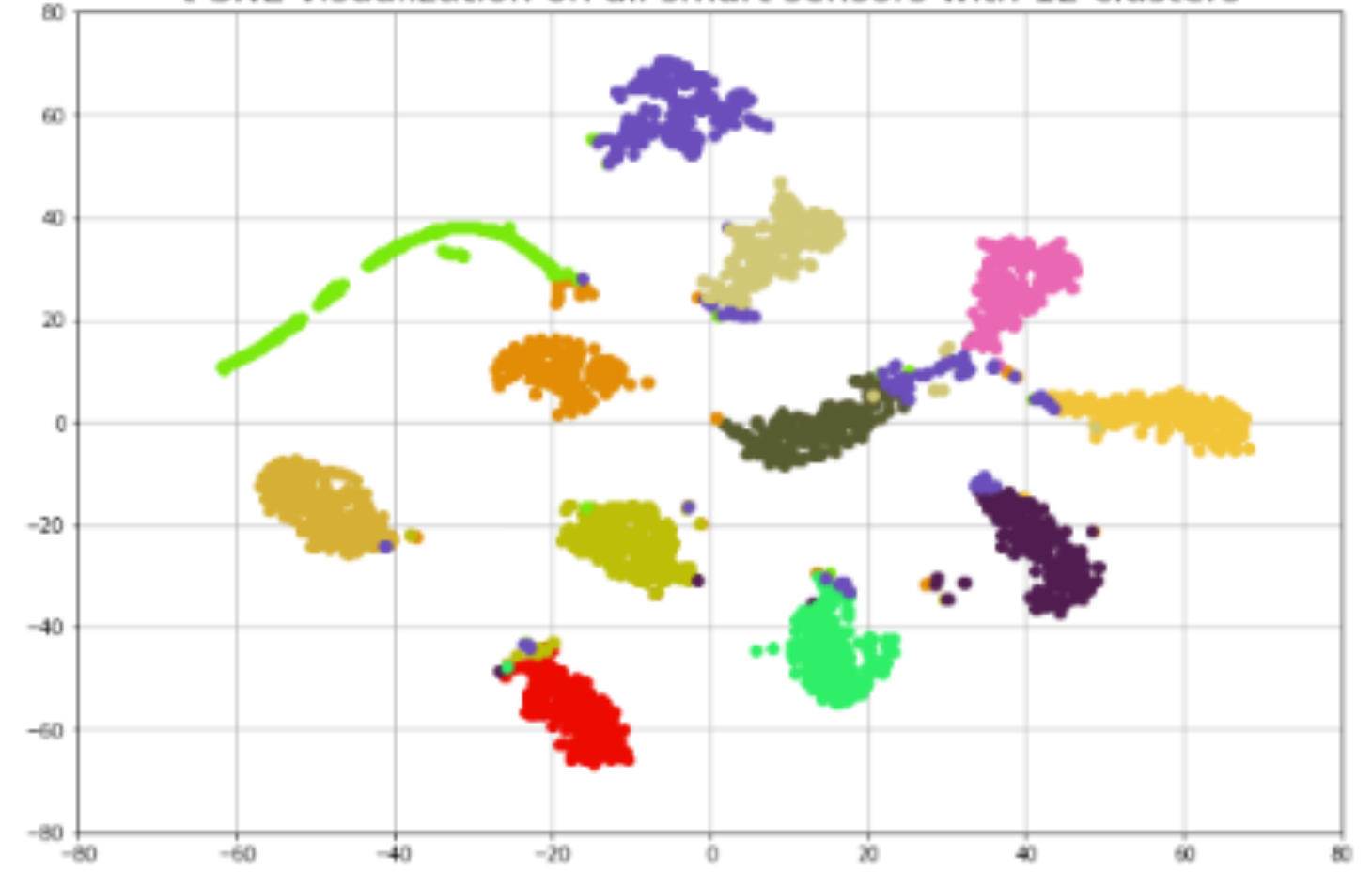}
    \caption{t-SNE visualisation on the min-max normalised month-level training data coloured by FCM clusters.}
    \label{fig:welongCluster}
\end{figure}

After performing daily aggregation, six features, including smart meter ID, consolidated consumption data, and time-related features such as ``day of week," ``day of month," and ``month," were utilised in the prediction stage. The resulting data was then subjected to a fuzzy c-means (FCM) clustering algorithm, with a predefined 12-cluster number. Fig. \ref{fig:welongCluster} depicts the resulting clusters. Upon examining the formed clusters, it was discovered that each cluster corresponded to a specific month. Rather than employing a single prediction model for the entire dataset, 12 distinct prediction models were developed, one for each cluster, that had a different monthly availability range but shared the same sensor pattern.

\begin{figure}[b]
    \centering
    \includegraphics[width=1\columnwidth]{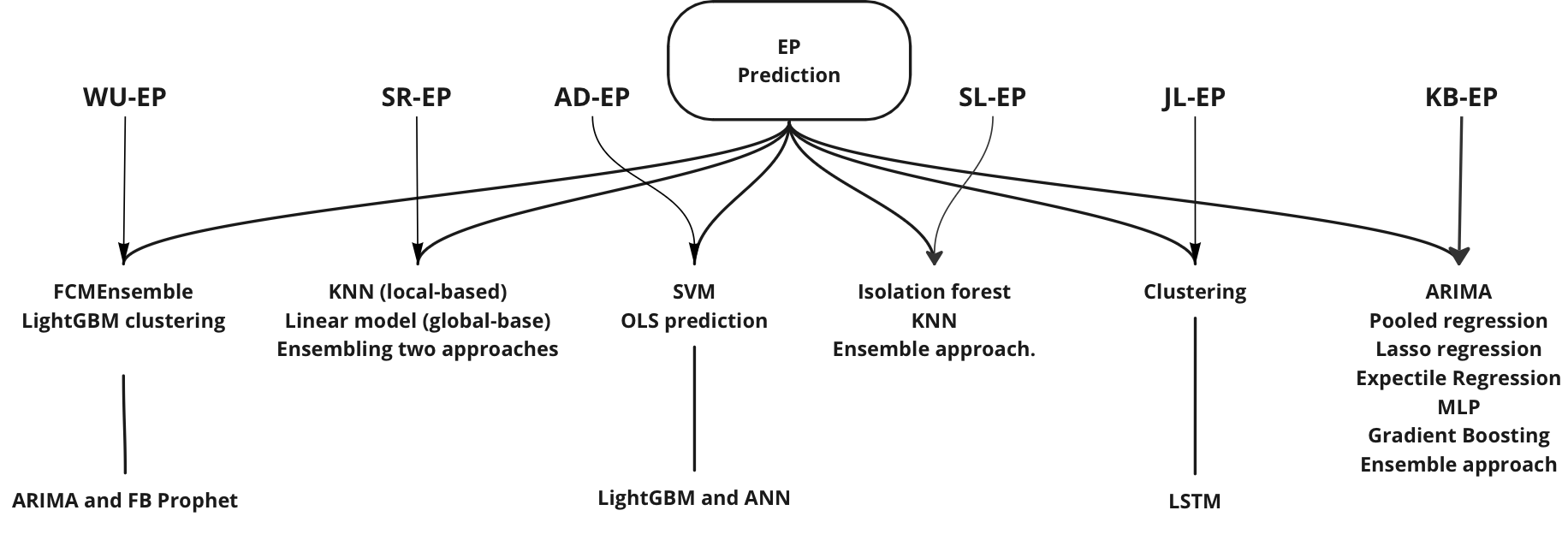}
    \caption{Prediction stage of the workflow, showcasing the EP strategies employed by prediction models and the other tried models if any.}
    \label{fig:TaxPredictionCIS}
\end{figure}

At the outset of the competition, the contestants experimented with various models such as Autoregressive Integrated Moving Average (ARIMA) \cite{hillmer1982arima} and Prophet \cite{taylor2018forecasting} on the provided dataset. However, the participants reported poor performance of these models, possibly due to the varying range of monthly availability of the data. For instance, households with meter readings starting in December led to statistical models failing to identify monthly trends over the course of a year, which in turn, resulted in the failure of data decomposition. Consequently, the Light Gradient Boosting Machine (LightGBM) \cite{lightGBM} model was utilised for the competition dataset.

The researchers used three LightGBM model structures to make predictions. The first prediction was made using a single LightGBM model on the whole dataset of 3248 households' consumption data. The second prediction was generated by creating 12 separate LightGBM models on the 12 clusters of households. The third prediction is implemented on the average sensor readings value of November and December from 2017 for each household and applied a LightGBM model to the dataset containing only the last two months of data.

During the post-processing phase, several steps were taken for each prediction. Firstly, households with smart meter readings that fell below a specified threshold for the last 31 days in 2017 were considered to have malfunctioning sensors, and a zero prediction was generated for them in 2018. Secondly, households with smart meter readings 20\% lower than the average had their latest month's readings used for prediction. Thirdly, a three-month window smoothing scheme was applied where the prediction for a particular month was based on the average energy consumption of the previous, current, and next month. Finally, for the winter season, the predictions were scaled up by a factor of 1.15, while for the summer season, they were scaled down by a factor of 0.85. The final prediction is emerged by ensemble the three structural predictions through the weighted average. The Python code of the approach can be accessed on: https://github.com/waylongo/cis-challenge-energy-prediction.

\subsubsection{SR-CIS} 




During the submission, two distinct approaches, namely the model-based and neighbour-based methods, were developed independently and subsequently ensembled. The model-based approach was grounded on the assumption that most meters share a common consumption trend that curves over the course of the year, utilising the exploratory data from the pre-processing step. Initially, a naive model was created by averaging the meter data per month, taking into consideration the heavily skewed consumption distribution by selecting Median over Mean. Additionally, the model employed the least square method for fitting, akin to linear regression, and implemented shifting and scaling techniques. On a daily level, the model leveraged correlation between meters and the K-Nearest Neighbors (KNN) approach to identify suitable substitute clusters. The fusion of these two distinct models, operating at both monthly and daily levels, culminated in the final result.

\subsubsection{AD-EP} 

The data that was aggregated monthly underwent a transformation process using the 
Box-Cox procedure \cite{ge1964box}. The time series were then divided into $12$ groups based on their availability, with Group $G_0$ having no missing observations and Group $G_{11}$ having only the December observation available. Group $G_0$ was chosen for further analysis, and the robust Singular Value Decomposition (SVD) 
\cite{candes2011robust} method was employed to represent each time series as a linear combination of 12 components, which were subsequently sorted by their influence based on the absolute value of their eigenvalues. 

Preliminary forecasting for the other groups was conducted accordingly.
As each group has varying numbers of observations for each smart meter, different sets of eigenvectors are utilised to create the preliminary forecast seasonal component for each group. An ordinary least square approach is employed to determine the coefficients for the principal components that provide the best approximation of the observed values. These coefficients are multiplied by the eigenvectors to generate the initial forecast results.

The initial step in the prediction phase involved calculating the residuals by subtracting the predicted values from the observed values where data was available. These residuals were then forecasted into the future using an exponential smoothing technique, which involved adding coefficients that increased exponentially and then reducing the sum by a factor  (both hyperparameters are ``manually" estimated). The resulting value was then added to the preliminary forecast to obtain the final prediction. The forecasted values were then transformed back to their original scale using the inverse Box-Cox transformation. 

Several approaches were initially considered, including the use of daily data and temperature, as well as  
LightGBM and neural network. However, due to time constraints, the investigation into these approaches and models was incomplete.

\subsubsection{SL-EP} 

To predict monthly consumption, the \textit{fraction} of each meter is first computed by dividing the monthly consumption by the total yearly consumption. The isolation forest \cite{liu2012isolation, pedregosa2011scikit} algorithm is then utilised to remove 
``abnormal" fractions, with 22 meters being identified as outliers and eliminated from the prediction. The remaining 248 meters are used as a base for forecasting, with a 
KNN approach applied. This involves calculating the Euclidean distance between meters and selected base meters and selecting \textit{k} base meters with the smallest distance. The prediction is generated by computing the 
Mean over the consumption distributions of the base meters. The base number \textit{k} is randomly chosen between 10 and 40, and the experiment is repeated five times to obtain an ensemble consumption prediction. 


\subsubsection{JL-EP} 

In the proposed prediction method, the consumption of each meter is estimated by utilising the clusters centroids obtained from pre-processing algorithms. The prediction is performed separately for each month, starting from the sign-up month. The consumption for a given month is estimated by finding the closest cluster centroid to the consumption of the meter and scaling up the corresponding distance by considering the future months' distances. This procedure is repeated for January to May
but not for June to December, where the re-scaling is omitted. 

An ensemble learning approach is also implemented, where the prediction is repeated assuming that the meter signed up in each of the following months, and the final prediction is based on the Median over all individual predictions.

For meters that signed up in December, the prediction is based on the consumption of the N closest neighbours that have a similar consumption in December, and the final prediction is the Median over these N meters. 

A low-pass filter is also applied to smooth the predicted consumption for each meter and month, which considers the average of the predicted month, the previous month, and the next month. The proposed approach was compared with a Long Short-Term Memory (LSTM) neural network, but the latter did not perform better in terms of accuracy.

   

\begin{figure}[b]
    \centering
    \includegraphics[width=0.65\columnwidth]{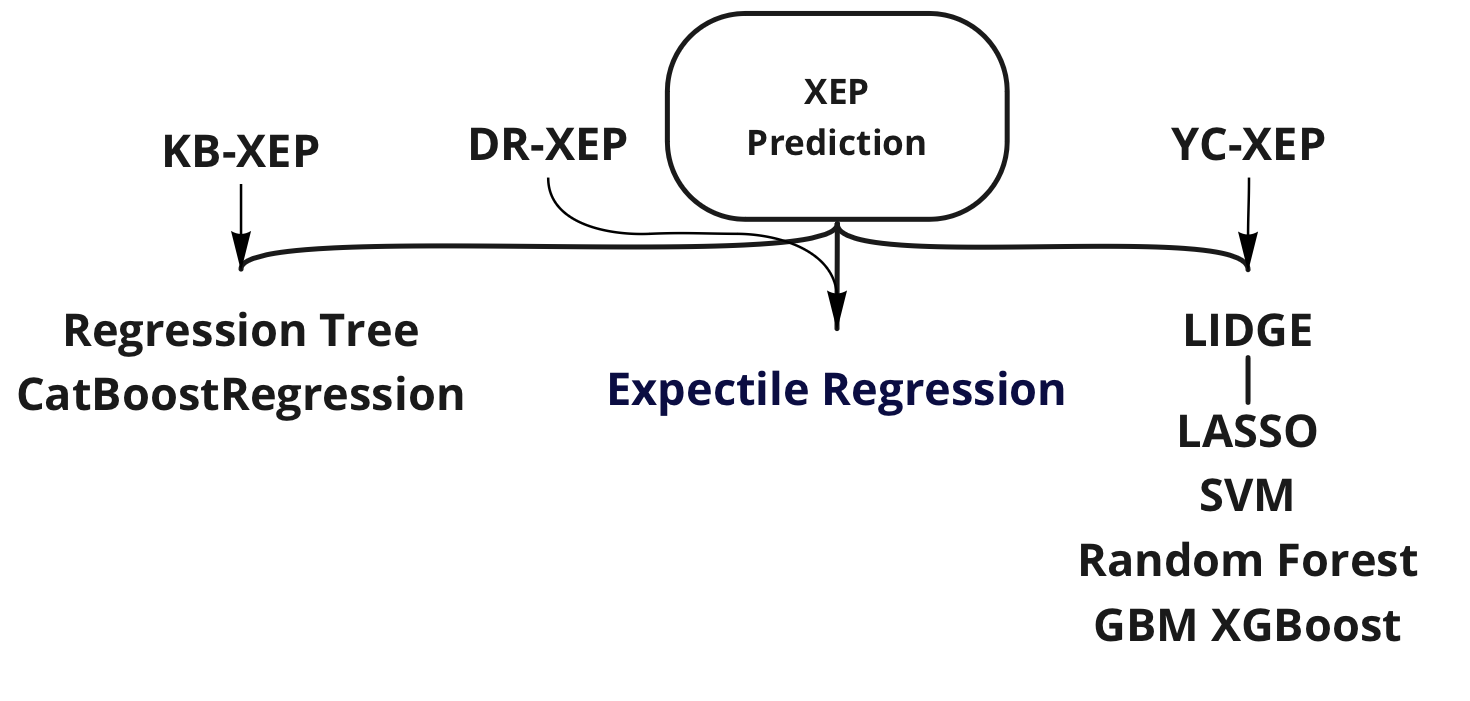}
    \caption{Prediction stage of the workflow, showcasing the XEP strategies employed by prediction models.}
    \label{fig:TaxPredictionFUZZ}
\end{figure}

\subsubsection{KB-EP}

The prediction employs an actual ensemble strategy with multiple models, as described below.

(i) The linear model ARIMA is applied on daily consumption data, and temperature data is used as an external variable. (ii) Pooled linear regression model is used with both consumption and temperature data pooled to create the lag matrix. This method is implemented by using the R package glmnet \cite{friedman2010regularization} (iii) Lasso linear regression \cite{tibshirani1996regression} model is integrated with the pooled regression to avoid over-fitting issues. This method is implemented by using the R package fForma \cite{montero2020fforma} (iv) Multilayer perceptron neural network non-linear model is implemented with the day of the week and month information as external variables. This method is implemented by using the R package nnet \cite{venables2013modern} (v) 
CatBoost without lags non-linear model is used with all external data. (vi) 
CatBoost with lags non-linear model is used with the day of the week, month, and past lags of 20 days as input variables. (vii) Random forest non-linear model is implemented using the R randomForest package~\cite{liaw2002classification}, with day of the week and month used as external variables.

After applying all the models on a daily basis, their daily forecasts for the year 2018 are aggregated to the monthly level. In the aggregation process, the geometric mean and median ensembling approaches are used to combine the individual forecasts.

\subsection{XEP Prediction Solutions}

\subsubsection{KB-XEP} 

In this methodology, the CatBoost regression tree model is employed as the main model for forecasting purposes. The model is capable of handling missing data (NAs) and takes into account all external variables, known as exogenous variables, to improve its predictive accuracy. This ensures that all available information is utilised to produce reliable forecasts.

\subsubsection{DR-XEP}

To predict daily consumption for the next day, an Expectile regression model is utilised, which is an extension of a conventional regression model \cite{newey1987asymmetric,schnabel2009optimal}. The model incorporates input features such as consumption over the previous 20 days, a calendar feature for the month, and a bootstrapped temperature obtained from past temperatures for the predicted day. The resulting one-day-ahead forecasts are recursively fed back into the model to generate forecasts iteratively for the whole year. The Dynamic Time Warping (DTW) technique is employed to find the nearest time series to the current one, followed by a bootstrapping procedure based on the seasonal-trend decomposition procedure based on loess (STL) decomposition and a moving block bootstrap from \cite{webb2001discovering} to bootstrap those series. To get a forecast for each of the bootstrapped series from the neighbourhood, the last window of 20 data points from that series is used as input. Finally, features such as mean, maximum, and minimum consumption over the window and mean temperature for the entire month are predicted.

\subsubsection{YC-XEP}
Multiple machine learning algorithms, including Lidge, Lasso, Support Vector Machine (SVM), Random Forest, 
Gradient Boosting Machine (GBM) and eXtreme Gradient Boosting (XGboost), are employed for power consumption prediction. However, due to inconsistent external data and missing values, the Lidge model 
overcomes the other models in terms of performance. The analysis also indicates that the power consumption of each meter varies with temperature, rather than time. To address missing values, the 2017 data is used as the outcome of the 2018 power consumption forecast.




\begin{table}[b]
\caption{EP overall prediction scores}
\label{tab:predictionCIS}
\begin{center}
\begin{tabular}{|l|l|l|l|}
\hline
\textbf{Contestant} & \textbf{Yearly RAE} & \textbf{Monthly RAE} & \textbf{Total RAE} \\ \hline
\textbf{WU-EP}     & 0.2864              & \textbf{1.0078}      & \textbf{0.6471}   \\ 
\hline
\textbf{SR-EP}     & \textbf{0.2825}     & 1.0486               & 0.6655             \\ \hline
\textbf{AD-EP}     & 0.2816              & 1.0564               & 0.669              \\ \hline
\textbf{SL-EP}     & 0.2875              & 1.0728               & 0.6801            \\
\hline
\textbf{JL-EP}     & 0.2892              & 1.0828               & 0.686             \\ 
\hline
\textbf{KB-EP}     & 0.3017              & 1.0754               & 0.6885             \\ \hline
\end{tabular}
\end{center}
\end{table}

\subsection{Overall Prediction Approaches}

In the EP competition, as demonstrated in Fig. \ref{fig:TaxPredictionCIS}, contestants used various approaches to predict household electricity consumption. 
WU-EP used 
an FCM clustering algorithm to group households based on their consumption patterns per month, developed 12 different models for each cluster, and then used an ensemble technique to get the final prediction. SR-EP developed two different methods, model-based and neighbour-based, that use the monthly average or median consumption of similar households, and then combined them in the final prediction. AD-EP used the robust SVD decomposition method to transform the data into a linear combination of 12 components and then used exponential smoothing to predict the residuals of the model. SL-EP used the isolation forest algorithm to identify outliers and eliminate them from the prediction. Overall, the approaches used machine learning and statistical techniques such as clustering, ensemble methods, and decomposition to forecast electricity consumption, with some of them focusing on outlier detection, seasonal decomposition, and time-series forecasting.

In the XEP competition, as demonstrated in Fig. \ref{fig:TaxPredictionFUZZ}, the three contestants used different approaches to predict power consumption. The KB-XEP used the CatBoost regression tree model, which can handle missing data and utilises all external variables. DR-XEP used an expectile regression model that incorporates various input features and employs DTW and bootstrapping procedures to generate iterative forecasts for the entire year. YC-XEP used multiple machine learning algorithms, and the Lidge model provided the best performance due to inconsistent external data and missing values. 
Overall, the contestants utilised different techniques, but all aimed to improve forecasting accuracy. The approaches were evaluated based on their ability to handle missing data, external variables, and produce accurate forecasts. 


Overall, after comparing the approaches from all contestants across two conference competitions (see reported prediction scores in Tables \ref{tab:predictionCIS} and \ref{tab:predictionFUZZ}), there is no preferred approach outperforming the others. However, it can be concluded from all the approaches that using sophisticated models such as DNN or LSTM does not necessarily lead to better performance, as the data and availability of external variables vary significantly throughout the year, and there is no consistency in registering meters. Therefore, contestants opted for simpler approaches that appeared to provide better performance for the given problem. It can be inferred that the simpler models, such as XGboost and regression-based models, are more suitable for the forecasting task given the complexity and variability of the data.

Regarding how solutions in Table \ref{tab:predictionFUZZ} are less accurate than solutions in Table \ref{tab:predictionCIS}, since in EP the goal was optimising predictions while in XEP competition aimed for a good balance between interpretability and accuracy, it becomes evident that the choice of competition objectives significantly influenced the strategies employed by the participants. While the EP competition focused primarily on optimising predictions, leading to more accurate results, the XEP competition emphasised the importance of interpretability alongside accuracy. Therefore, contestants in the XEP competition, while achieving a balance between interpretability and accuracy, might have sacrificed some predictive performance in comparison to their counterparts in the EP competition. This highlights the intricate trade-offs that arise when addressing multifaceted challenges in energy consumption prediction.


\begin{table}[]
\caption{XEP overall prediction scores}
\label{tab:predictionFUZZ}
\begin{center}
\begin{tabular}{|l|l|l|l|}
\hline
\textbf{Contestant}   & \textbf{Yearly RAE} & \textbf{Monthly RAE} & \textbf{Total RAE} \\ \hline
\textbf{KB-XEP}      & 0.3333     & \textbf{1.4062}               & \textbf{0.8697  }           \\ \hline
\textbf{DR-XEP}      & \textbf{0.3303}              & 1.5683               & 0.9493             \\ \hline
\textbf{YC-XEP} & 0.3537              & 1.6879               & 1.0208             \\ \hline
\end{tabular}
\end{center}
\end{table}

\section{Interpretability}

\begin{figure}[b]
    \centering
\includegraphics[width=0.45\columnwidth]{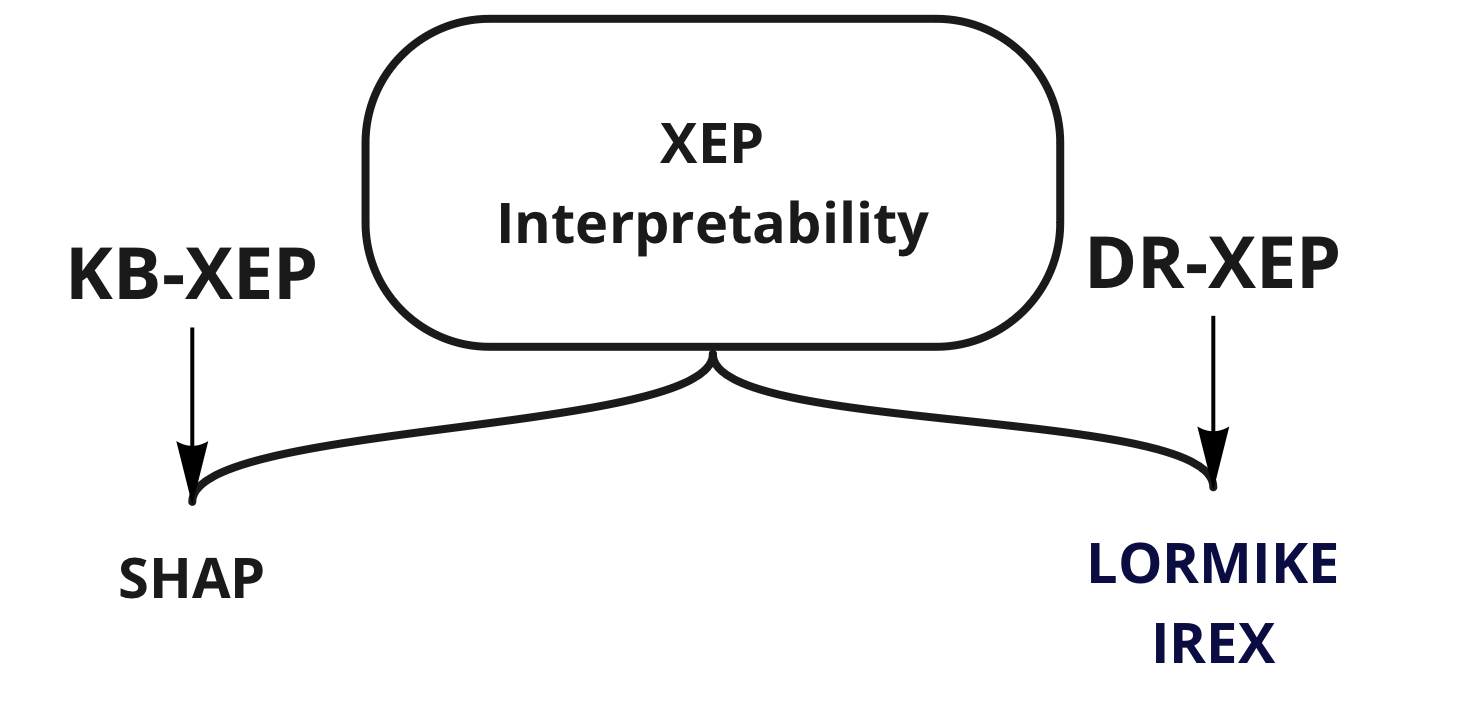}
    \caption{Strategies adopted for maximising interpretability in the XEP competition.}
   \label{fig:taxInterpret}
\end{figure}

In this section, we will delve into the solutions for maximizing model interpretability that were proposed by the shortlisted participants in the XEP competition (see a visual summary in Fig. \ref{fig:taxInterpret}). As pointed out in Fig. \ref{fig:workflow}, we will provide a detailed overview of the proposed methods for ensuring transparency by imposing interpretability constraints, handling the interpretability-accuracy tradeoff, as well as generating linguistic explanations.

In addition, as mentioned above, for the XEP conference competition, a dedicated website was developed to evaluate the selected methods, where yearly and monthly predictions for ten smart meters were displayed to experts, along with corresponding explanations and actual data. The evaluation was conducted anonymously, and experts ranked the 
predictions along with the given explanations, for all the 10 selected smart meters, in terms of the 10 
quality criteria Ci, thus allowing for a focused and comprehensive assessment of the methods' performance and interpretability. 

Let us describe below first the proposed solutions (Section \ref{sec:int:finalists}) and second their overall assessment (Section \ref{sec:int:overall}).

\subsection{Interpretability Approaches}
\label{sec:int:finalists}

\subsubsection{KB-XEP} 

This contestant opted for a local post-hoc interpretability analysis supported by the SHapley Additive exPlanation (SHAP) \cite{lundberg2017unified} method, which is one 
of the most well-known methods in the context of 
XAI.
The rationale behind SHAP is supported by the game theory method proposed by Lloyd Shapley in 1953.
SHAP explains individual predictions using the coalitional game’s best Shapley values.
Such Shapley values yield information about the weighted total contribution of all potential feature values which are taken into account by a given model.

In the yearly forecast explanation, the 
3 most relevant attributes, according to SHAP, are chosen 
in agreement with user requirement, 
to fill in the following template-based textual explanation: 

\begin{displayquote} The estimation of your energy consumption for the next year is mostly influenced by the following attributes: [\textit{`SHAP-F1', `SHAP-F2', `SHAP-F3'}].
\end{displayquote}

Accordingly, an example of explanation as an instantiation of this template is as follows:

\begin{displayquote} The estimation of your energy consumption for the next year is mostly influenced by the following attributes: [\textit{`month', `max-temp', `num-bedrooms'}].
\end{displayquote}

In addition,
some actionable 
complementary pieces of linguistic information are generated and added to the previous explanation.
To do so, some additional features, which are deemed as actionable by experts, are selected even if their Shapley values were smaller, to fill in additional templates.
An example of this complementary information is as follows:

\begin{displayquote} Your consumption may reduce by controlling the following devices: [\textit{`tv', `pc', `tumble dryer'}].
\end{displayquote}


In the explanation of the monthly forecast, akin to the annual protocol, the top \textit{3} attributes of the monthly forecast are emphasised. Furthermore, a comparative narrative with respect to the prior month is incorporated in the text exposition. Contingent upon the availability of data, certain descriptive terms (e.g., much lower or slightly higher) are established by thresholds and integrated into the operational description, as illustrated in the
following example:

\begin{displayquote} In September, your energy consumption will be \textbf{much lower} because of the following attributes: [ \textit{`max temp', `num bedrooms', `month'}]. Your consumption may reduce by controlling the following devices and what is related to them: [\textit{`tv', `pc', `set top box'}].
\end{displayquote}

\begin{table}[b]
\caption{XEP Interpretability scores}
\label{tab:interpretabilityScores}
\resizebox{\columnwidth}{!}{\begin{tabular}{|c|c|c|c|c|c|c|c|c|c|c|}
\hline
\textbf{Contestants}  & \textbf{C1}   & \textbf{C2}  & \textbf{C3}   & \textbf{C4}   & \textbf{C5}   & \textbf{C6}  & \textbf{C7}   & \textbf{C8}   & \textbf{C9}   & \textbf{C10}  \\ \hline
\textbf{DR-XEP}      & \textbf{3.82} & \textbf{3.9} & \textbf{3.97} & \textbf{3.95} & \textbf{3.92} & \textbf{3.8} & \textbf{3.67} & \textbf{3.37} & \textbf{3.57} & \textbf{3.97} \\ \hline
\textbf{KB-XEP}      & 3.45          & 3.22         & 3.05          & 3.22          & 3.17          & 3.1          & 3.37          & 2.5           & 3.27          & 3.12          \\ \hline
\textbf{YC-XEP} & 3.4           & 1.77         & 1.67          & 1.67          & 1.75          & 1.75         & 1.65          & 1.87          & 2.9           & 1.72          \\ \hline
\end{tabular}}
\end{table}

\subsubsection{DR-XEP}

A novel explainer has been developed with the aim of generating understandable explanations. This 
explainer is based on the LoRMIkA classification algorithm 
\cite{rajapaksha2020lormika}, which was previously developed for explaining machine learning classifiers and is employed here as a surrogate approach with association rules to achieve local interpretability. The method is arguably more comprehensible than other local surrogate models because the rules do not attempt to cover the entire feature space and can instead focus on the more interesting portions of the space, including both data instances and features. 
In addition to factual explanations that support the current prediction and counterfactuals that refer to why other specific predictions are not supported \cite{alonso2021s}, the new explainer also provides two additional types of rules known as ``current contradicting rules" and ``hypothetically supporting rules". 
Such rules highlight which aspects of a data instance are in disagreement with the current prediction and which features could be modified to further reinforce the prediction.

The methodology 
combines the LoRMIkA explainable classifier with the IREX impact rule mining algorithm \cite{webb2001discovering}, which is suitable for numeric outputs such as regression as opposed to classification. One key distinction is that impact rules can identify segments of a population that might have a higher or lower average value for the predicted variable. 


Given a specific smart meter, the LoRMIka+IREX algorithm makes the prediction and identifies the rules which support and explain such prediction.
Then, linguistic explanations are generated by filling in predefined templates where rule premises and conclusions are translated into natural language. 
Below you can see an illustrative example:

    

\begin{itemize}
\item Current supporting rules:
\begin{displayquote}
 Your predicted consumption is 119.02kWh, this is supported by your target month being February. 
 \end{displayquote}

 \item Current Contradicting Rules:
\begin{displayquote}
 The conditions that currently exist that indicate a risk of increased consumption by 179.55kWh for the particular month are 6.05 $<$ temperature $\leq$ 6.31.
\end{displayquote}

\item Hypothetically Supporting Rules:
\begin{displayquote}
 The conditions that need to be satisfied to maintain the monthly predicted consumption would be temperature $>$ 6.31.
 \end{displayquote}

\item Hypothetically Contradicting Rules (Counterfactual Rules):
\begin{displayquote}
 If you have a mean consumption $>$ 13.99 it may increase your consumption by 388.48kWh.
\end{displayquote}

\end{itemize}

\subsubsection{YC-XEP}
The team opted to concentrate on other aspects of energy consumption prediction and did not include any specific solution regarding interpretability issues.
On the contrary, this approach prioritised different facets of the problem. 




\subsection{Overall Interpretability Assessment}
\label{sec:int:overall}


The two explanation methods presented in this study adopt a model-agnostic approach, which attempts to generate template-based linguistic explanations 
for the given predictions. 
On the one hand, the well-known SHAP approach
appears to be more targeted towards specific predictions, and it selected 
the three 
most relevant features for both monthly and yearly prediction explanations.
Template-based linguistic explanations combined information provided by SHAP with linguistic expressions such as ``slightly lower". 
While this approach can provide targeted predictions, it also provides clear actionable guidance to consumers. 
On the other hand,
the combination of LorMIKA and IREX algorithms, intending to scrutinize the extracted rules and identify any contradictions, 
proved to be quite useful in providing clear direction to customers.

As it can be seen in Tables \ref{tab:interpretabilityScores} and \ref{tab:fuzzScores}, DR-XEP was the winner.
It was also the best regarding Yearly RAE (see Table \ref{tab:predictionFUZZ}).
Even if KB-XEP achieved higher accuracy regarding Monthly RAE and Total RAE, it was clearly worse regarding interpretability, as can be deduced from the lower global assessment (see C10 in Table \ref{tab:interpretabilityScores}).
Both contestants, DR-XEP and KB-XEP, got scores from the expert panel, in average above 3 (with Likert scales from 1 to 5) for all the ten criteria.
The biggest difference appears in C8, that is the only criteria for which KB-XEP is below 3.

\begin{table}[t]
\centering
\caption{XEP Overall Contestant Scores}
\label{tab:fuzzScores}
\begin{tabular}{|l|l|}
\hline
Contestants & Score \\ \hline
DR-XEP     & 7.84  \\ \hline
KB-XEP     & 7.38  \\ \hline
YC-XEP       & 4.82  \\ \hline
\end{tabular}
\end{table}

\section{Beyond the Competitions}

In this section, we will discuss the aspects beyond the given dataset and competitions, exploring potential avenues for further research and development in the field of energy consumption prediction. While the competitions focused on improving accuracy and interpretability, there are additional areas that can be addressed to enhance the utility and impact of such dataset and models.

\subsubsection{Data Enrichment} Firstly, while the provided dataset offers valuable insights, there is untapped potential in leveraging external data sources to enrich the prediction patterns. For instance, incorporating additional data such as household types or utilities information could provide a more comprehensive understanding of energy consumption factors and contribute to more accurate predictions \cite{yildiz2017recent}. Integrating external data could enhance the models' ability to capture complex relationships and improve overall forecasting performance.

\subsubsection{Energy Disaggregation} Moreover, once there is a deeper comprehension of the enriched dataset, scaling up the analysis to larger datasets becomes a viable option. This scalability can facilitate the exploration of energy disaggregation techniques, which involve isolating the individual appliance-level energy consumption from the overall household consumptions \cite{kolter2011redd}. By leveraging the patterns and insights extracted from the enriched dataset, researchers can apply machine learning techniques to identify and track energy usage at the appliance level. This fine-grained information can provide valuable insights for energy management initiatives, enabling households to understand the energy consumption of individual appliances and make informed choices regarding their usage. 

\subsubsection{Anomaly Detection} At the household level, pattern detection plays a pivotal role in understanding and optimising energy consumption. An example of such pattern detection can be the development of an alarm system to identify anomalies in energy consumption patterns at household levels. By analysing historical data and establishing baseline patterns, the system can identify significant deviations in energy usage, which may indicate unusual events or even emergencies. For instance, drastic changes in energy usage by elderly households, compared to their normal usage patterns, could raise concerns and trigger further investigation or intervention. Leveraging household-level patterns and implementing such alarm systems empowers energy providers and support services to proactively address potential issues, provide timely assistance, and ensure the safety and well-being of households.

\subsubsection{Response Programme} The detection of overall consumption patterns within households can also be highly valuable in the implementation of demand response programmes \cite{chen2018measures}. By leveraging machine learning algorithms to analyse energy consumption patterns, these programmes can incentivise households to adjust their energy usage based on grid conditions and energy availability. For instance, by shifting their energy consumption to off-peak hours or aligning it with periods of high renewable energy generation, households can actively contribute to load balancing and promote grid stability. One specific application of this approach can be the detection of electric vehicles (EVs) within households. By tracking the energy consumption patterns associated with EV charging, demand response programmes can optimise EV charging schedules to align with renewable energy peaks or periods of lower grid demand. This integration of EV detection within demand response programmes showcases how the overall detection of consumption patterns can enhance energy management strategies and extend to various applications, beyond just EVs, to effectively balance energy demand and maximise the utilisation of renewable energy sources.

\subsubsection{``Comprehensive" Understanding} Once the interpretability of the models is evaluated, energy-driven factors can be communicated with households to optimise their billing and promote energy-efficient behaviours \cite{alonso2018}. By providing households with clear explanations of the factors influencing their energy consumption, researchers can design personalised energy reports or interfaces that highlight the impact of various factors, such as weather conditions, occupancy, or appliance usage, on their energy bills. This transparency empowers households to make informed decisions regarding their energy usage and identify opportunities for energy-saving practices. Additionally, behavioural interventions, such as providing tailored energy-saving tips or suggesting energy-efficient appliances, can be integrated into these interfaces to further encourage sustainable behaviours and facilitate energy-conscious choices.

\subsubsection{Responsible AI} Addressing concerns regarding responsible AI and data privacy is essential in these endeavours \cite{arrieta2020explainable,alonso2023,dignum2019}. As researchers explore the potential of utilising smart meter data and external sources, ethical considerations must guide the development and deployment of energy management systems. Safeguarding data privacy and ensuring data security should be prioritised, with robust measures in place to protect individuals' personal information and prevent unauthorised access or misuse of data. Transparent data governance frameworks and compliance with regulatory guidelines are crucial for maintaining public trust and confidence in the responsible use of data. 
Responsible AI practices should be integrated throughout the entire process, from data collection and pre-processing to model development and deployment, to ensure the ethical and socially responsible use of AI in energy management applications.

By addressing these aspects beyond the competitions and dataset, researchers can leverage the insights gained to propel advancements in energy consumption prediction, management practices, and customer engagement. These areas present valuable opportunities to optimise energy utilisation, enhance sustainability, and foster a more efficient and informed energy ecosystem. The exploration of external data integration, energy disaggregation, pattern detection, demand response programmes, personalised communication with households, and responsible AI considerations collectively contribute to the development of innovative and effective energy management strategies that benefit both individuals and the larger energy system.
















\section{Conclusion}

To summarise, this paper has highlighted the significance of accurate and interpretable energy consumption forecasting, particularly at the household level. The findings from the EP 2020 and XEP 2021 competitions have provided valuable insights into addressing the challenges associated with energy prediction using real-world smart meter data. The proposed solutions have showcased the effectiveness of various pre-processing, prediction, and 
XAI approaches in improving the accuracy and interpretability of energy consumption forecasts. However, it is important to acknowledge that real-world datasets present inherent challenges, such as inconsistencies, missing values, seasonal variations, and limited availability of data. These challenges must be carefully considered when developing machine learning models for energy prediction. Nonetheless, the competitions have provided a platform for researchers to collaborate and contribute to the development of innovative solutions that have the potential to revolutionise the energy industry and promote sustainable energy consumption practices indirectly.

Furthermore, the competitions laid the groundwork for future research by establishing a benchmark problem for evaluating the performance of energy consumption prediction models and promoting the integration of advanced machine learning techniques and alternative data sources. The findings extended beyond the competitions, encouraging further exploration of energy disaggregation, demand response programs, and behaviour interventions. Ultimately, these competitions provide an invaluable dataset for residential energy research and contribute significantly to fostering a sustainable and efficient energy ecosystem. As the field progresses, researchers and practitioners are poised to leverage the valuable insights gained from these competitions to develop more transparent, accurate, and interpretable models, thereby contributing to a greener and more sustainable future.

\section*{Acknowledgment}
The authors would like to thank the support of the IEEE-CIS Task Force on Explainable Fuzzy Systems when running the XEP 2021 competition. I. Triguero is funded by a Maria Zambrano Senior Fellowship at the University of Granada.



%

\bibliographystyle{unsrt}  
\bibliography{references}

\end{document}